\documentclass[twocolumn,superscriptaddress,preprintnumbers,amsmath,amssymb]{revtex4-1}
\usepackage{graphicx}
\usepackage{dcolumn}
\usepackage{bm}
\usepackage[normalem]{ulem}
\usepackage[usenames,dvipsnames]{xcolor}

\begin{document}
\preprint{ \color{Blue} DRAFT v1.6 - \today}

\title{\large \color{Blue} Brainbots as smart autonomous active particles with programmable motion}

\author{M.~Noirhomme}
\affiliation{GRASP, Institute of Physics B5a, University of Li\`ege, B4000 Li\`ege, Belgium.}
\author{I.~Mammadli}
\affiliation{PULS, Institute for Theoretical Physics, FAU Erlangen-Nürnberg, 91058, Erlangen, Germany}
\author{N.~Vanesse}
\affiliation{GRASP, Institute of Physics B5a, University of Li\`ege, B4000 Li\`ege, Belgium.}
\author{J.~Pande}
\affiliation{Department of Physical and Natural Sciences, FLAME University, Pune, India.}
\author{A.-S.~Smith}
\affiliation{PULS, Institute for Theoretical Physics, FAU Erlangen-Nürnberg, 91058, Erlangen, Germany}
\author{N.~Vandewalle}
\affiliation{GRASP, Institute of Physics B5a, University of Li\`ege, B4000 Li\`ege, Belgium.}
\maketitle


{\bf 

We present an innovative robotic device designed to provide controlled motion for studying active matter. Motion is driven by an internal vibrator powered by a small rechargeable battery. The system integrates acoustic and magnetic sensors along with a programmable microcontroller. Unlike conventional vibrobots, the motor induces horizontal vibrations, resulting in cycloidal trajectories that have been characterized and optimized. Portions of these orbits can be utilized to create specific motion patterns. As a proof of concept, we demonstrate how this versatile system can be exploited to develop active particles with varying dynamics, ranging from ballistic motion to run-and-tumble diffusive behavior.

}

\vskip 4 mm
\hrule 
\vskip 1 mm


\section{Introduction}

Active matter consists of numerous active particles, each of which consumes energy to generate motion or exert a specific force. Consequently, these systems are inherently out of thermal equilibrium \,\cite{MSAM}. Unlike thermal systems relaxing towards equilibrium and systems with boundary conditions imposing steady currents, active matter systems break time-reversal symmetry because energy is continually dissipated by the individual constituents. Many examples of active matter are seen in biological systems and span all the scales of living organisms, from bacteria\,\cite{DCMBCC, FMBSOATDCRC} to fish schools\,\cite{EESDTFS, SSPCMZ}. Pioneering studies have evidenced the complex behaviors of these living systems: group formation\,\cite{HGDPF, CM}, swarming and flocking\,\cite{SSMPDDDFSM, CMSSAG, CVBDPUCCAGM}, and transport in complex landscapes \cite{CTMPSFB}, to name a few.

\begin{figure}[th]
	\centering
 (a)   \includegraphics[width=0.75 \columnwidth]{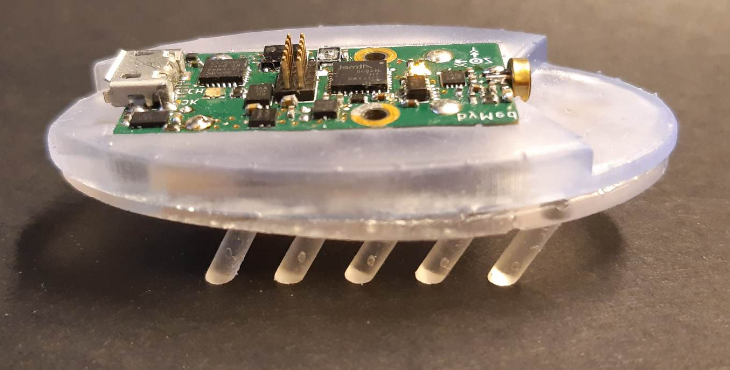}\\
 (b)   \includegraphics[width=0.75 \columnwidth]{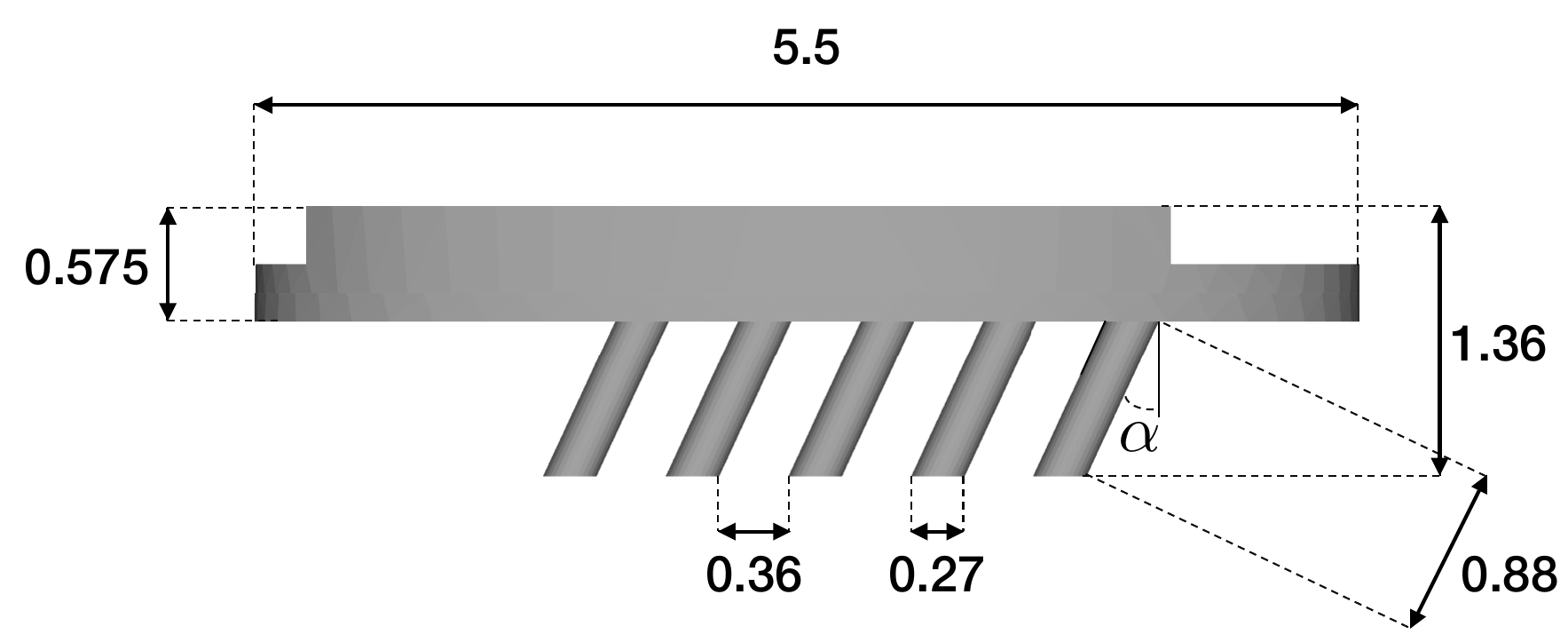}\\
	\caption{(a) Picture of the autonomous brainbot. The elliptic body is $5.5\,$cm long and $3\,$cm wide. Electronic components are placed on top. (b) Side view of the 3D-printed part of the brainbot with specific lengths in centimeters. 
 }\label{fig:Brainbot}
\end{figure}

Recent efforts have been directed towards developing synthetic self-propelled particles at different scales. Magnetic or Janus colloids are able to self-assemble into moving arrangements \cite{palacci}, mimicking bacteria and other tiny organisms. Mesoscopic machines can be created along liquid interfaces, imitating the locomotion of mesoscopic organisms like copepods \cite{hubert} and allowing the exploration of swimming at intermediate Reynolds numbers. Asymmetric inert particles of a few millimeters in length can be excited by a vibrated plate to induce horizontal collective motions leading to vortex formation, and phase separation in crowded environments \cite{CMVPD, LROVPD, VPDSMBCCD, RRMCSO}. They are often called vibrobots and their main characteristic is their design of inclined legs, important for transferring vertical vibrations into horizontal motion.

Based on a similar vibration-driven locomotion, various commercial robotic particles are available such as hexbugs \cite{DSPPHT, SCAAS} and kilobots \cite{KLCRSODCB}. Hexbugs are equipped with at least one motor with a horizontal axis, inducing vertical vibrations. Their motion has been studied in harmonic traps \cite{DSPPHT}, and in elastic arrays in which they exhibit synchronized motion of crystal-like structures \cite{SCAAS}. They have also been used in realizing thermodynamic concepts like the Szil{\' a}rd engine \cite{Roichman2023}, and in experiments where they change their complex environments to create paths \cite{Roichman2024}. A few vibrobots have been placed in a single elastic membrane for studying the motility of a single cell passing through an aperture \cite{Kellay}. Multiple synthetic cells have been placed in a confined space for creating a synthetic confluent tissue \cite{arora}. This robotic class of synthetic active particles is being studied more and more these days because it is cost effective and allows the design and execution of human-scale experiments. 

The key challenge concerning these robotic particles is the control of their motion \cite{epje2024}. Vibrobots primarily exhibit ballistic behavior. For instance, hexbugs undergo straight trajectories until they collide with other particles or the boundaries of the system. However, it can be very useful to have robotic systems capable of executing other kinds of motion, such as diffusive motion. For example, Brownian-like motion can help in testing physical laws, and run-and-tumble motion, which is intermediate between Brownian and ballistic motion \cite{Bechinger}, can be valuable for mimicking bacteria and other living organisms. In such motions, reorientation of the particle is essential. Therefore, an autonomous device capable of following controlled motion that includes both translation and rotation would be highly desirable.

In this article, we address this issue by creating active, programmable vibrobot particles. The main idea is to change the vibrator orientation -- and, more precisely, the motor axis -- from horizontal to vertical. Horizontal vibrations are expected to induce reorientation of the device. An example of such a bot, which we call a brainbot, is illustrated in Fig.~\ref{fig:Brainbot}. 

Section~\ref{sec2} presents the technical characteristics of the brainbot. A deep understanding of its multifaceted motion is necessary before it can be employed to execute a desired specific motion. Section~\ref{sec3} presents a complete study of its locomotion and then explains how to build a specific trajectory, from a ballistic path to a Brownian one.

\section{Autonomous brainbots}\label{sec2}

\subsection{Body}

A typical 3D-printed body of a brainbot is shown in Fig.~\ref{fig:Brainbot}. In the current paper the brainbots have five pairs of inclined legs, though it is possible to change their number as well as their tilt angle (denoted by $\alpha$). The material to 3D-print the body is usually rigid ABS (acrylonitrile butadiene styrene), but flexible resin seems to provide enough friction to optimize the motion. The shape of the 3D-printed body can be modified for creating other kinds of active particles like square or disk particles allowing for specific assemblies when they collide or pack together. In the present paper, the elliptical shape is fixed.

The leg geometry and the elasticity of the material are key ingredients for optimizing the locomotion. In the next subsections, the motion of the robot has been characterized for different tilt angles $\alpha$ between $5^\circ$ and $25^\circ$. 

\subsection{Motor}

The device is a centimeter-sized structure propelled by vibrations induced by a motor placed inside the particle. The latter is made of an asymmetric mass rotating in the device along the horizontal plane. Its geometry is cylindrical and it has a diameter of 10 mm and a thickness of 2.7 mm, and a mass of approximately 1 g. It is placed at the front of the robot, while the battery (the heaviest component) is placed at the rear, so as to direct the movement of the particle forward. The motor rotating the asymmetric mass is subjected to a 3V potential difference of the PWM (Pulse Width Modulation) type, through the modification of an effective voltage $V_\text{E}$. Accordingly, the rotational speed of the motor can be varied from 0 to 11000 rpm.

When the engine is started from a state of rest, it does not reach its maximum rotation speed instantly. In the same way, when the vibrator is running and the motor is braked, there is another delay before it comes back to rest. As explained in Section~\ref{sec3}, these delays have a direct impact on the optimal locomotion of the brainbot. 

\subsection{Electronic components}

The motor and the sensors are all connected to a programmable micro-controller placed on top of the device as shown in Fig.~1(a). Because it is fully programmable, we have named our device the brainbot. 

\begin{figure}
	\centering
	\includegraphics[width=0.95 \columnwidth]{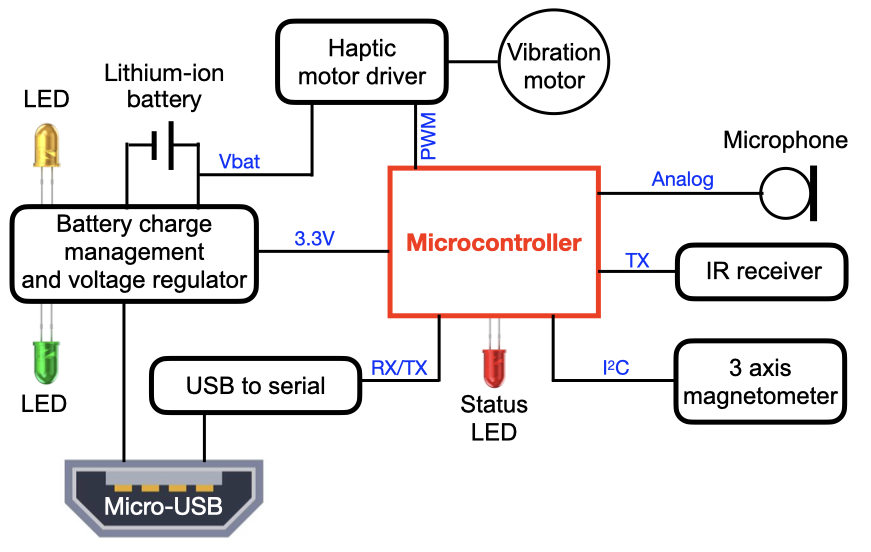}
	\caption{Simplified circuit diagram of the brainbot electronics.}\label{fig:CircuitDiagram}
\end{figure}

Fig.~\ref{fig:CircuitDiagram} shows a simplified circuit diagram of the brainbot electronics. The central part consists of an $8$ bits microcontroller running a dedicated firmware. The system is powered by a button lithium-ion rechargeable battery. The battery starts charging when the brainbot is connected to a micro-USB port. A circuit manages the charging process and also protects the battery against deep discharge. In the basic version of the brainbot, three sensors are connected to the microcontroller: 
\begin{itemize}
    \item An Infrared (IR) receiver is used to get a signal from a remote control. A common TV remote control can be used to trigger the particle activity.
    \item A directional microphone is placed on the front of the brainbot to analyze the ambient noise. It can, for example, detect the noise produced by other neighboring brainbots.
    \item A $3$-axis Hall magnetometer is placed on the device to measure the magnetic field strength and orientation. Brainbots can orient along the magnetic field of the Earth or move in a magnetic landscape.
\end{itemize}
Other sensors can be added in the near future such as light sensors, accelerometers, and inclinometers. The brainbots are programmed in C like for the usual Arduino platform. In this way, several behaviors can be implemented for piloting the brainbot, such as remote-controlled movement, random displacement, and location tracking.

\subsection{Main characteristics}

In Tab.~\ref{tab:example} we provide the main technical characteristics of the brainbots. Some of these characteristics can be modified by 3D-printing other bodies or extended by adding different sensors. 

\begin{table}[h]
\caption{\label{tab:example}Technical characteristics of a brainbot sorted into different categories.}
\begin{ruledtabular}
\begin{tabular}{lll}
Bulk & Mass & $13\,$g \\ \hline
Size    &  Length & $5.5\,$cm \\
        &  Width  & $3\,$cm \\
        &  Height & $1.5\,$cm \\ \hline
Legs    & Number  & $2 \times 5$ \\
        & Inclination $\alpha$ & $5^\circ - 25^\circ$ \\ \hline
Battery & Autonomy & $60\,$min \\
        & Charge duration & $80\,$min \\ \hline
Magnetometer & Sensitivity threshold & $50\,$mT  \\ \hline
Microphone  & Sensitivity & $-42$ dB \\
            & Bandwidth & $100-10000$ Hz \\ \hline
Motor   & PWM               & $50\%-100\%$ \\
        & Effective voltage $V_\text{E}$ & $0-3$ V \\
        & Rotational speed  & $0-11000$ rpm
\end{tabular}
\end{ruledtabular}
\end{table}

\begin{figure}[h]
	\centering
	\includegraphics[width=0.95 \columnwidth]{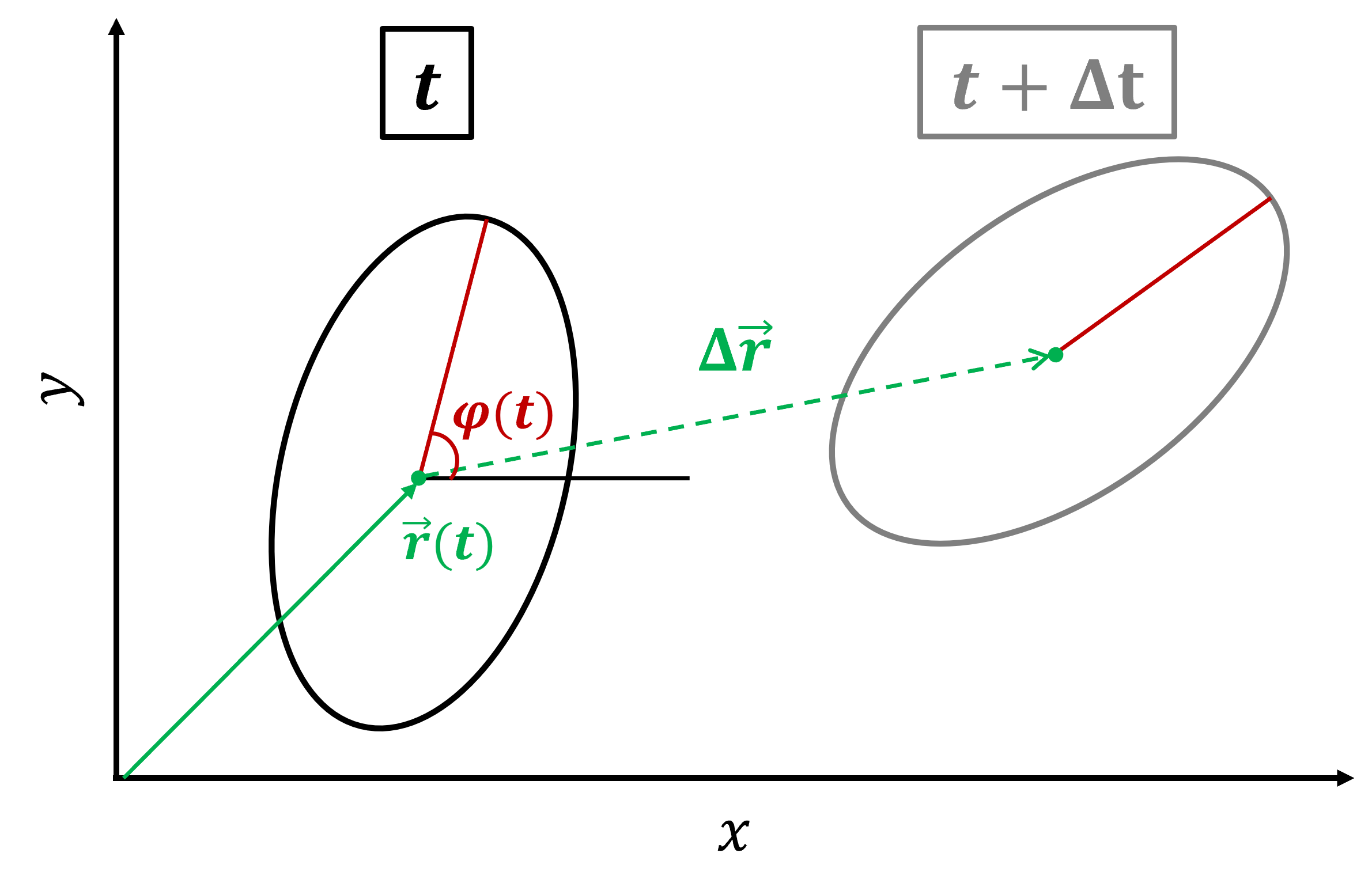}
	\caption{The elliptical robot in planar motion. At time $t$, $\vec r(t)$ is the position vector of the geometrical centre and $\varphi(t)$ is the 
    angle of orientation between the $x$-axis and the major axis of the ellipse.}\label{fig:drdt}
\end{figure}

The brainbot is placed in a rectangular arena on a planar and horizontal surface. A camera placed above tracks the brainbot position $\vec r$ and orientation $\varphi$ by using a homemade OpenCV routine. The position $\vec r$ corresponds to the geometrical centre of the elliptical particle and is different from the centre of mass. The orientation is deduced from the semi-major axis angle $\varphi$ (Figure \ref{fig:drdt}).

\section{Locomotion}\label{sec3}

\subsection{Characterization of spontaneous trajectories} \label{subsec3A}

\begin{figure*}
	\centering
	\includegraphics[width=2\columnwidth]{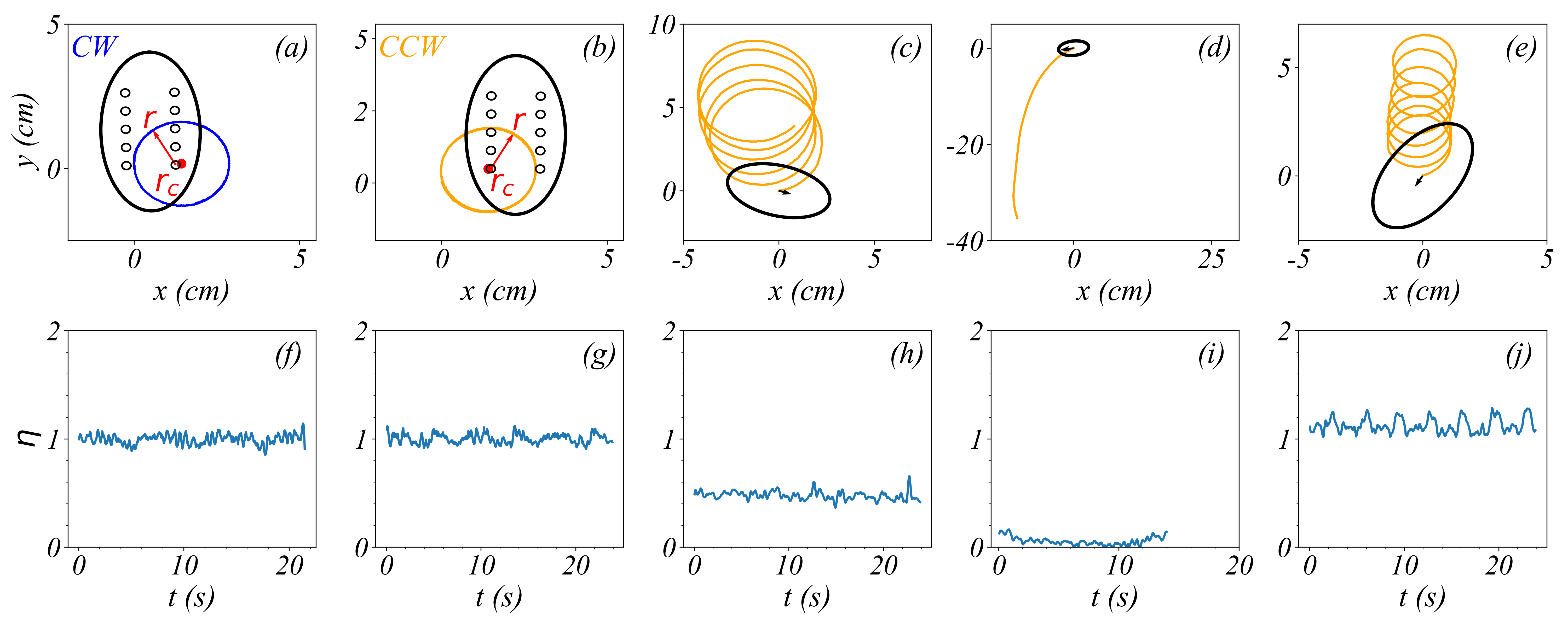}
	\caption{Brainbot trajectories (upper panels) with their respective $\eta$ variation (lower panels). In the upper panels, the black ellipses indicate the initial position and orientation of the bots, with the small black circles within the ellipses indicating the positions of the legs. The brainbot ellipses are drawn to scale relative to the ensuing trajectories of the centre of mass, marked in blue in panel (a) (for clockwise spinning) and orange in panels (b) to (e) (for counterclockwise spinning). Furthermore, in panels (a) and (b) the red dots indicate the instantaneous centres of rotation of the spinning motion, and the red arrows indicate the vector $\vec r$ - $\vec r_\text{c}$.}\label{fig:eta}
\end{figure*}

Motion of the brainbot is achieved by triggering the internal motor, whose power, and the resulting intensity of the vibrations, can be controlled through the effective voltage $V_\text{E}$. Under constant $V_\text{E}$ we observe four main kinds of trajectories (orange curves in Fig.~\ref{fig:eta}, upper panels). A brainbot may spin around a point in its interior while remaining otherwise fixed in position (Fig.~\ref{fig:eta}(a)-(b)), it may spin around a point in its interior while also executing translatory motion (Fig.~\ref{fig:eta}(c)), it may exhibit purely translatory motion without spinning (Fig.~\ref{fig:eta}(d)), or it may move backwards (Fig.~\ref{fig:eta}(e)). The spinning component of the motion may be in either the clockwise (Fig.~\ref{fig:eta}(a)) or the counterclockwise sense (Fig.~\ref{fig:eta}(b)). The particular kind of motion executed, as well as the speed of the brainbot and the radius of the circular component, appears to depend on the voltage $V_\text{E}$ as well as the inclination angle $\alpha$ (Fig.~\ref{fig:Brainbot}) of the legs of the robot. 

To characterize the different kinds of motion systematically, we first consider the cases where the brainbot spins around a point in its interior, which happens for low values of the effective voltage $V_\text{E}$ of the motor. We observe that in these cases, the centre of rotation does not coincide with the geometric centre of the object, but appears to lie close to one of the rear legs. This is not surprising as the heavier battery is in the rear part, and because the only contact with the substrate is through the legs. This is similar to the case with ``differential-drive-like brushbots" in \cite{DIFFDRIVE}, though the bots there are circular rather than elliptical.

We use the Savitzky-Golay \cite{savgol} filter (Supplementary Material) to clean the trajectories for noise and then calculate the velocity and the angular velocity with numerical derivatives (the trajectory and angle data are sampled at sufficiently high frequencies to allow such derivatives), 
\begin{equation}
\begin{aligned}
    \vec v = \frac{d\vec r}{dt} \quad  \text{and} \quad \omega = \frac{d \varphi}{dt}.
\end{aligned}
\end{equation}
Then the well-known relationship between the two velocities, when the position vector of the instantaneous centre of rotation is $\vec r_\text{c}$, is
\begin{equation}
     \vec v = \vec \omega \times (\vec r - \vec r_\text{c}),
     \label{eq:icr}
 \end{equation}
where $\vec v = (v_x,\;\; v_y,\;\; 0)^T$  and $\vec \omega = (0,\;\;0,\;\;\omega)^T$ are the velocity and the angular velocity vector, respectively. The analysis is limited to the two-dimensional $x-y$ plane. 

For purely spinning motion, the centre of rotation $\vec r_\text{c}$ is fixed. Since we know $\vec r$, $\vec v$ and $\vec \omega$ at any given time from the observed trajectories, we can calculate $\vec r_\text{c}$ from Eq.~(\ref{eq:icr}). Figs.~\ref{fig:eta}(a) and \ref{fig:eta}(b) show $\vec r_\text{c}$ for clockwise and counterclockwise rotational motion, respectively. The results confirm the idea that for a purely (or predominantly) spinning motion, the centre of rotation (marked as red dots in Figs.~\ref{fig:eta}(a) and \ref{fig:eta}(b)) falls on one of the rearmost legs. The distance between the geometric and rotational centres for purely spinning motion is thus
\begin{equation}
    |\vec r-\vec r_\text{c}| = l_{\text{leg}} = 1.45 \text{ cm.}
\end{equation}

To characterize the other trajectories, we define a new parameter, $\eta$, as
\begin{equation}
    \eta = \frac{|\omega| \  l_{\text{leg}}}{|\vec v|}.
    \label{eq:eta}
\end{equation}
In the general case, the instantaneous centre of rotation is not necessarily on a rear leg and can lie inside or outside the bot. 

The parameter $\eta$ is a quantitative measure of how similar the observed motion is to spinning. For purely spinning motion, the trajectories are circles and $\eta$ should be $1$ by definition. For purely translatory (linear) motion without any spinning, $\eta$ should be $0$, because $\omega$ should be $0$ in this case (as $\varphi$ should be constant). For $\eta$ values between $0$ and $1$, a combination of spinning and translatory motion may be expected.

The lower panels in Fig.~\ref{fig:eta} demonstrate precisely this behavior of $\eta$ corresponding to the different kinds of trajectories. Its value, which fluctuates over time for a given trajectory, generally falls between $0$ and $2$. Values close to $1$ indicate predominantly spinning motion (Fig.~\ref{fig:eta}(f)-(g)), values between $0$ and $1$, which occur the most often in experiments, indicate a combination of spinning and linear translatory motion (Fig.~\ref{fig:eta}(h)), and values close to 0 indicate pure translation without spinning (Fig.~\ref{fig:eta}(i)). In addition, we also observe rare trajectories for which $\eta > 1$ (Fig.~\ref{fig:eta}(j)), indicating that the centre of rotation in these cases is closer to the geometrical centre than the rear legs. In these cases the bot typically moves backward.

\begin{figure}
    \centering
\includegraphics[height=0.348\textwidth]{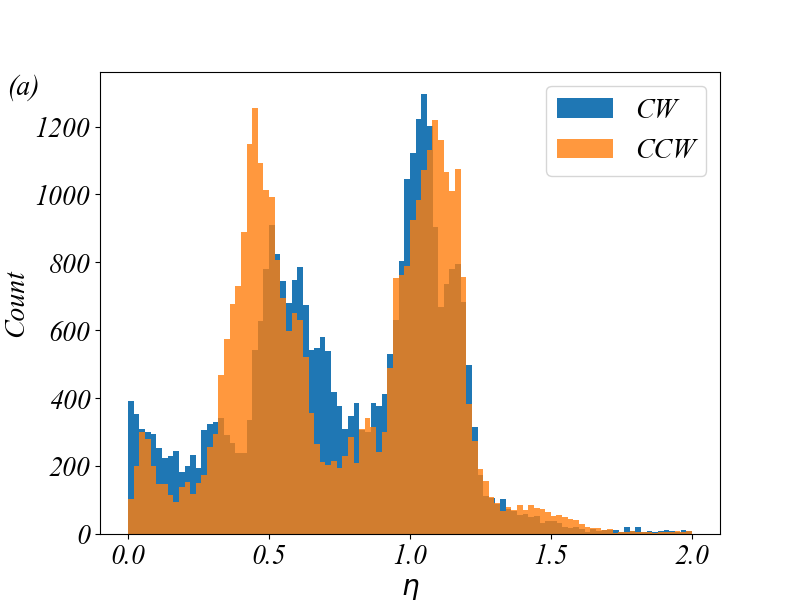} 
\includegraphics[height=0.348\textwidth]{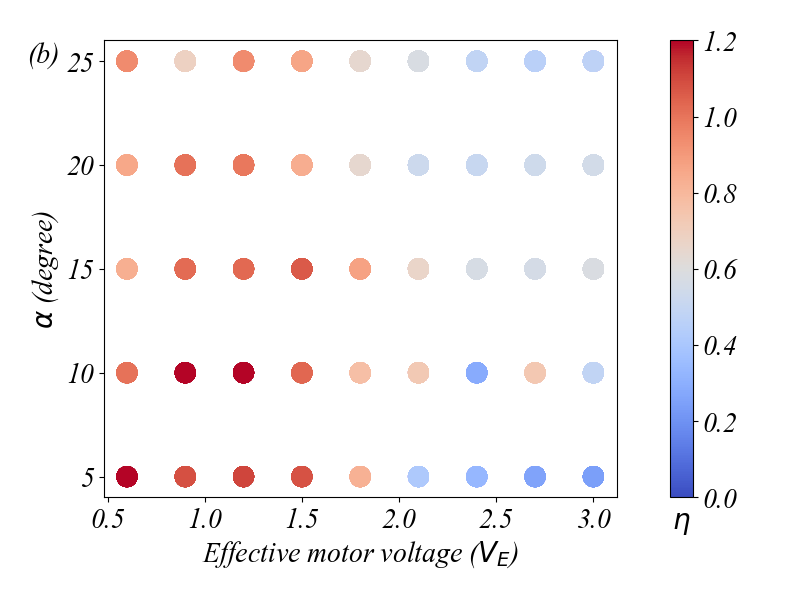}
    \caption{(a) Frequency distribution of $\eta$ values across all experiments (in which the effective motor voltage $V_\text{E}$ and the leg angle $\alpha$ are varied). The distribution is similar for both clockwise and counterclockwise spins, and is dominated by $\eta \simeq 1.0$ and $\eta \simeq 0.5$, corresponding to pure spinning and a combination of spinning and translation, respectively. (b) Dependence of $\eta$ on the leg angle $\alpha$ and the effective motor voltage $V_\text{E}$. For low motor voltage, $\eta$ has values around $1$, indicating a motion dominated by spinning. For higher motor voltages, $\eta$ becomes smaller and smaller in value, indicating that the translatory component of the motion becomes more and more prominent. For small $\alpha$ values the motion tends to be dominated by either spinning or translation, and when $\alpha$ increases combinations of spinning and translation become more prominent.}\label{fig:etaphase}
\end{figure}

We find that the frequency distribution of all $\eta$ values from all experiments (Fig.~\ref{fig:etaphase}(a)) is similar for both clockwise (blue) and counterclockwise (orange) spins. In both cases the most frequent modes are those of pure spin ($\eta=1$) and a combination of spinning and translatory motion ($0<\eta<1)$. The third most common mode is that of pure translation without spinning ($\eta\rightarrow0$), but it is observed much less frequently than the other two.

Since the parameter $\eta$ efficiently characterizes the different kinds of trajectories observed in the experiments, we can capture the effect of the effective motor voltage $V_\text{E}$ and the leg inclination angle $\alpha$ on the motion of the brainbot through their effect on $\eta$ (Fig.~\ref{fig:etaphase}(b)). As expected, for low motor voltages the motion is dominated by spin (corresponding to $\eta$ values close to $1$), while at higher voltages translation becomes prominent (with $\eta$ close to $0$). As the leg angle $\alpha$ increases, there is a tendency for the motion to be a combination of the spinning and translation modes, whereas for low $\alpha$ values one or the other of these modes, depending on the effective voltage $V_\text{E}$, tends to dominate.

\subsection{Encoding ballistic motion} \label{Ball}
By appropriately stitching together the spontaneous trajectories described in Section~\ref{subsec3A}, we can build specific kinds of motion of our brainbots. In order to obtain an efficient straight trajectory, successive instances of clockwise and counterclockwise rotations can be performed. Fig.~\ref{fig:Random}(a) shows two independent trajectories formed by these alternating rotations, in which brainbots zigzag along an overall single direction. Suppose $T$ denotes the duration of each rotation and $\alpha$ denotes the angle between the chords defining two successive rotations (where these chords are formed by joining the initial and final points of the curve in a single rotation, as shown in Fig.~\ref{fig:Random}(a)). Then, one can show that the translational mean speed $\bar{v}$ of the robot is
\begin{equation}\label{eq:meanspeed}
\bar{v} = \frac{2 R}{T} \sin{\left(\frac{\omega T}{2} \right)} \sin{\left(\frac{\alpha}{2} \right)} \mathrm{,}
\end{equation}
where $\omega$ is the rotational speed and $R$ is the radius of each circular trajectory. This speed should be considered as an upper limit since at each rotation change, there is some delay before the brainbot reaches the maximum angular speed value. A careful observation of the trajectories of Fig.~\ref{fig:Random}(a) reveals such a behavior since the very short part of each arc has a variable radius of curvature $R$. We obtain the optimal mean translational speed $\bar{v} \approx 3.5$ cm/s when the duration of each sequence is $T \approx 1.5$ s.

\begin{figure}
	\centering
\includegraphics[height=0.33\textwidth]{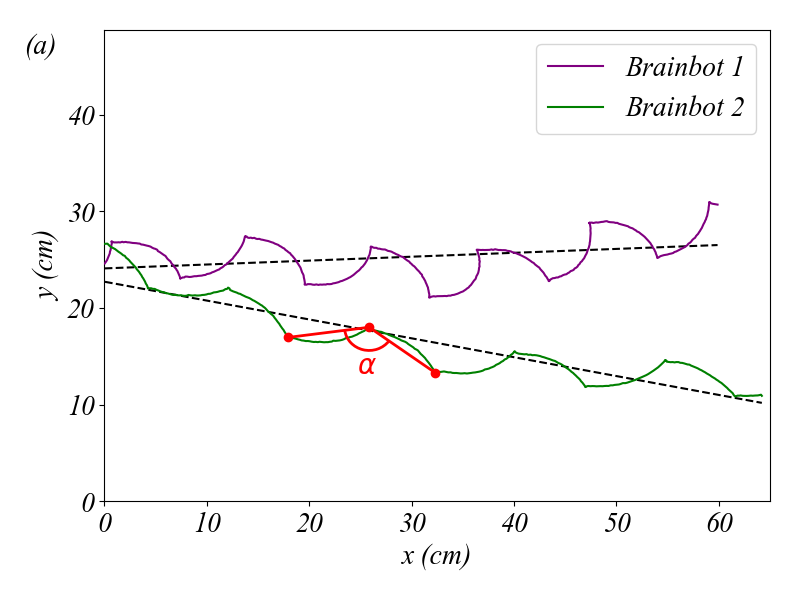}\label{fig:ballistic}
\includegraphics[height=0.33\textwidth]{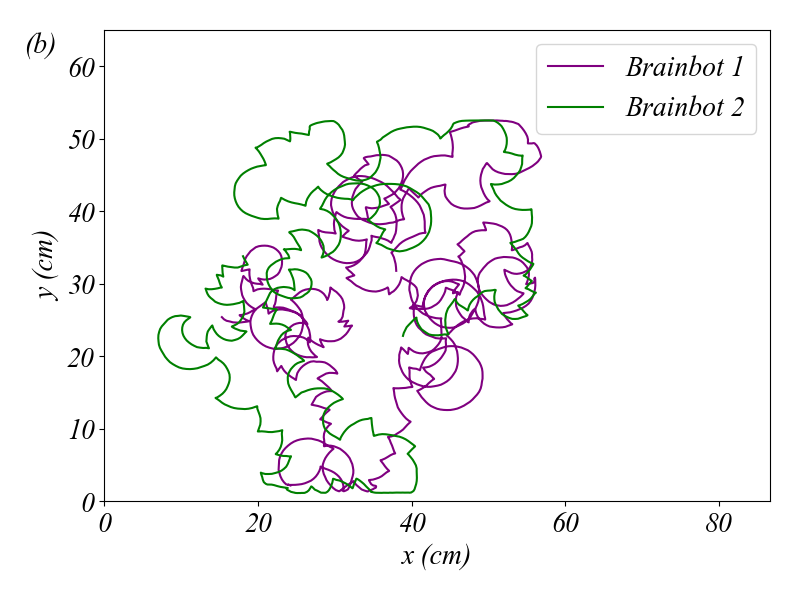}\label{fig:Random}
\includegraphics[height=0.33\textwidth]{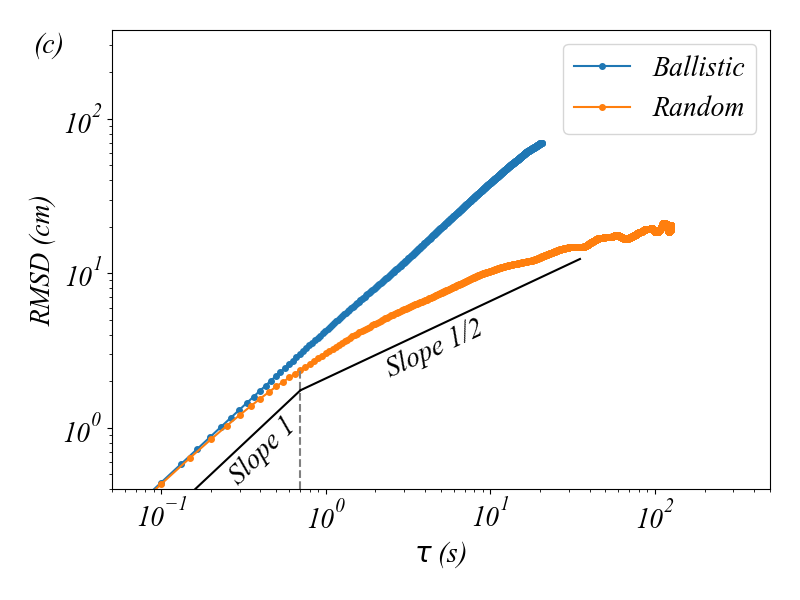}
	\caption{(a) Two independent trajectories obtained by applying sequences of clockwise and counterclockwise rotations of the asymmetric mass in the internal motor of a brainbot. Dashed lines correspond to approximate ballistic motions. (b) Two independent trajectories obtained by randomizing the application of both clockwise and counterclockwise rotations of the asymmetric mass in the internal motor. (c) Root Mean Square Displacement (RMSD) of a brainbot undergoing random motion (orange) or a nearly straight trajectory (blue). The black curve corresponds to a double linear fit, showing a separation between a rectilinear-motion-type regime (of slope 1 on the logarithmic scale) and a diffusive-type regime (of slope 1/2). The separation time scale is approximately $\tau^\star \approx 0.9$ s. }\label{fig:Random}
\end{figure}

Therefore, ballistic motion can be achieved when the brainbot performs alternating sequences of clockwise and counterclockwise rotations, and it is possible to control the translational speed by adjusting the duration $T$ of these sequences.

\subsection{Encoding diffusive motion}\label{Rand}

We can also make our bots undergo Brownian motion by randomizing the sequences of the spontaneous trajectories. At each step, we program the brainbot to randomly select the rotational direction (clockwise or counterclockwise), with a duration $T$ uniformly drawn from the interval $[0.4, 1.0]$ seconds. Since two consecutive sequences may at times adopt the same rotational direction, long circular arcs can occasionally appear. As illustrated in Fig.~\ref{fig:Random}(b), where two independent trajectories are represented, this type of motion appears diffusive, with trajectories filling the entire accessible space. To confirm this observation, we measured the Root Mean Square Displacement (RMSD) of the brainbot, defined by
\begin{equation}
\text{RMSD}= \sqrt{ \langle [\vec r(t+\tau) - \vec r(t)]^2 \rangle } \rm{,}
\end{equation}
where the angular brackets denote an average over time $t$ for fixed $\tau$ values. A linear behavior, $\text{RMSD} \sim \tau$, is expected for ballistic motion while a square root behavior, $\text{RMSD} \sim \tau^{1/2}$, is expected for diffusive motion. A review of the RMSD method to characterize dynamical behaviors in active matter can be found in \cite{Bechinger}.

Fig.~\ref{fig:Random}(c) demonstrates the RMSD of our brainbots, in a log-log plot, for two cases: ballistic motion as described in Section~\ref{Ball}, and random motion as defined above. To obtain the second curve, we conducted $12$ independent experiments of random motion. A straight line with a slope of $1$ is observed for ballistic motion, whereas the random motion exhibits linear behavior at short time scales followed by a square root dependence at longer time scales. The diffusive behavior is limited at long times due to the finite size of the arena. The characteristic time separating these behaviors is approximately $\tau^\star \approx 0.9$ s, corresponding to the duration required to decorrelate the alternating ballistic sequences. 

This combination of linear and diffusive behaviors characterizes the run-and-tumble dynamics commonly observed in bacteria and other active systems \cite{Bechinger}. These dynamics serve as models for understanding how individual motion gives rise to collective patterns in biological systems. Programming run-and-tumble behavior in autonomous vibrobots represents a significant advancement as such systems are typically limited to ballistic motion. This advancement will, in the future, enable novel experiments to explore fundamental physical laws in active matter, particularly in systems where fluctuations are critical to the overall behavior.
\vskip 0.2 cm
\section{Conclusion}
In summary, we have developed a versatile and cost-effective robotic platform that can be used to build specific active systems. Our brainbot is fully programmable, and by thoroughly studying its dynamics, we identified the optimal parameters that increase its speed up to 3.5 cm/s. The natural motion tends to be either spinning or a combination of spinning and translation, and we can combine these kinds of motion appropriately to create a wide range of trajectories, from ballistic to diffusive. Diffusive trajectories are characterized by tunable run-and-tumble dynamics, adaptable for various applications. This programmability provides a unique platform for testing physical laws and exploring recent discoveries in active matter, such as phase separation and fluid-solid transitions.

In future work, we will focus on developing interactions between brainbots using their sensors. We also plan to comprehensively explore the interactions between brainbots and various soft and hard obstacles.

\vskip 0.2 cm
\section*{Acknowledgments} 

M.~Noirhomme thanks the Belgian Federal Science Policy Office (BELSPO) for the provision of financial support in the framework of the PRODEX Programme of the European Space Agency (ESA) under contract number 4000103267. The authors thank FNRS for financial support through Grant number PDR T.0251.20. Special thanks to M.~Mélard for his technical support for building brainbots. L.~Brennenraedts, N.~Martin, and S.~Noel are acknowledged for their involvement in this project. 

\vskip 0.2 cm
\section*{Supplementary material}
\begin{figure}[h]
	\centering
	\includegraphics[height=0.467\textwidth]{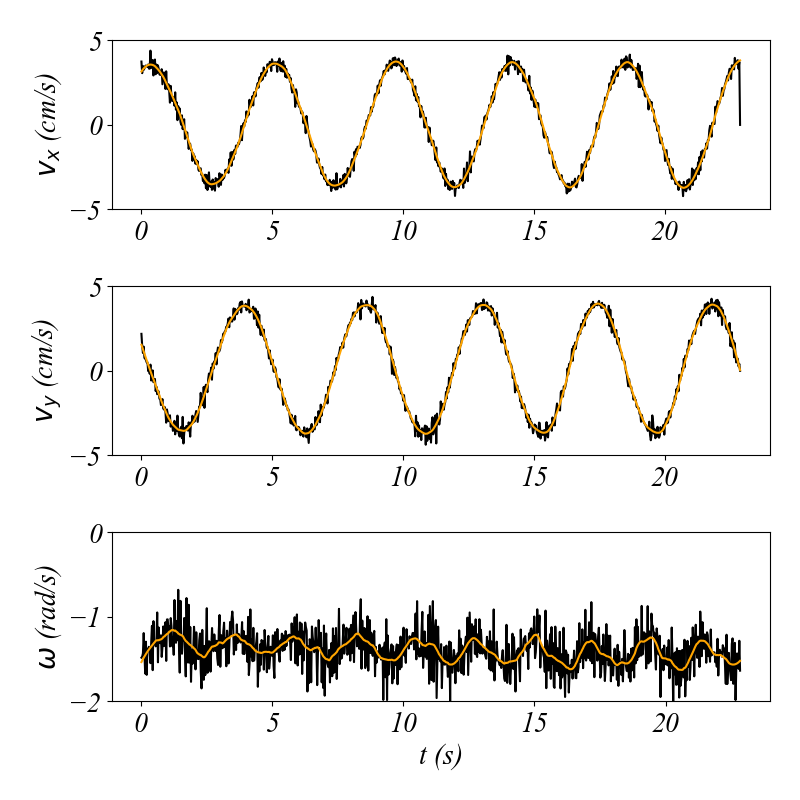}
	\caption{The black curves show the velocities ($V_x$, $V_y$ and $\omega$) when the relevant derivatives were computed directly from the raw data. The orange lines indicate the corresponding derivatives from data filtered using the Savitzky-Golay filter.}
    \label{fig:REFINED_TRAJ}
\end{figure}

Using trajectories in their raw forms results in noisy outputs. We cleaned the noise from the trajectories using the Savitzky-Golay filter \cite{savgol} and obtained smooth derivatives as shown in Fig.~\ref{fig:REFINED_TRAJ}. The Savitzky-Golay filter is a convolution-type filter that uses polynomial fits in given window sizes. The user is required to define the window size and the degree of the polynomial, which results in different levels of accuracy and smoothness. This filter is readily available in many packages, so its details are omitted here.


\end{document}